\def\year{2017}\relax
\documentclass[letterpaper]{article}
\usepackage{aaai17}
\usepackage{times}
\usepackage{helvet}
\usepackage{courier}
\usepackage{amsmath}
\usepackage{comment}
\usepackage{graphicx}
\usepackage{url}
\usepackage{color}
\usepackage{xspace}
\usepackage{dsfont}
\usepackage{algorithm}
\usepackage{subfigure}
\usepackage{algpseudocode}
\newcommand{\DSP}{Descriptive Space Partitioning\xspace}
\newcommand{\partitions}{partitions\xspace}

\newcommand{\partition}{partition\xspace}

\newcommand{\DSPshort}{DSP\xspace}
\newcommand{\UUB}{Bandit for Unknown Unknowns\xspace}
\newcommand{\UUBshort}{UUB\xspace}

\DeclareMathOperator*{\argmax}{arg\,max}

\nocopyright
\begin{document}
\title{Identifying Unknown Unknowns in the Open World: \\  Representations and Policies for Guided Exploration}
\author{Himabindu Lakkaraju$^{*}$, Ece Kamar$^{+}$, Rich Caruana$^{+}$, Eric Horvitz$^{+}$\\
$^{*}$Stanford University, $^{+}$Microsoft Research\\
$^{*}$himalv@cs.stanford.edu, $^{+}$ $\{$eckamar, rcaruana, horvitz$\}$@microsoft.com
}
\maketitle
\begin{abstract}
Predictive models deployed in the real world may assign incorrect labels to instances with high confidence. Such errors or \emph{unknown unknowns}  are rooted in model incompleteness, and typically arise because of the mismatch between training data and the cases encountered at test time. As the models are blind to such errors, input from an oracle is needed to identify these failures. In this paper, we formulate and address the problem of informed discovery of unknown unknowns of any given predictive model where unknown unknowns occur due to systematic biases in the training data.
We propose a model-agnostic methodology which uses feedback from an oracle to both identify unknown unknowns and to intelligently guide the discovery. We employ a two-phase approach which first organizes the data into multiple \partitions based on the feature similarity of instances and the confidence scores assigned by the predictive model, and then utilizes an explore-exploit strategy for discovering unknown unknowns across these \partitions. We demonstrate the efficacy of our framework by varying the underlying causes of unknown unknowns across various applications. To the best of our knowledge, this paper presents the first algorithmic approach to the problem of discovering unknown unknowns of predictive models.
\end{abstract}

\section{Introduction}
Predictive models are widely employed in a variety of domains ranging from judiciary and health care to autonomous driving. As we increasingly rely on these models for high-stakes decisions, identifying and characterizing their unexpected failures in the open world is critical. We categorize errors of a predictive model as: \emph{known unknowns} and \emph{unknown unknowns}~\cite{btm}. Known unknowns are those data points for which the model makes low confidence predictions and errs. On the other hand, unknown unknowns correspond to those points for which the model is highly confident about its predictions but is actually wrong. Since the model lacks awareness of its unknown unknowns, approaches developed for addressing known unknowns (e.g., active learning~\cite{settles2010active}) cannot be used for discovering unknown unknowns.  

\begin{figure}
\centering
\includegraphics[scale=0.15]{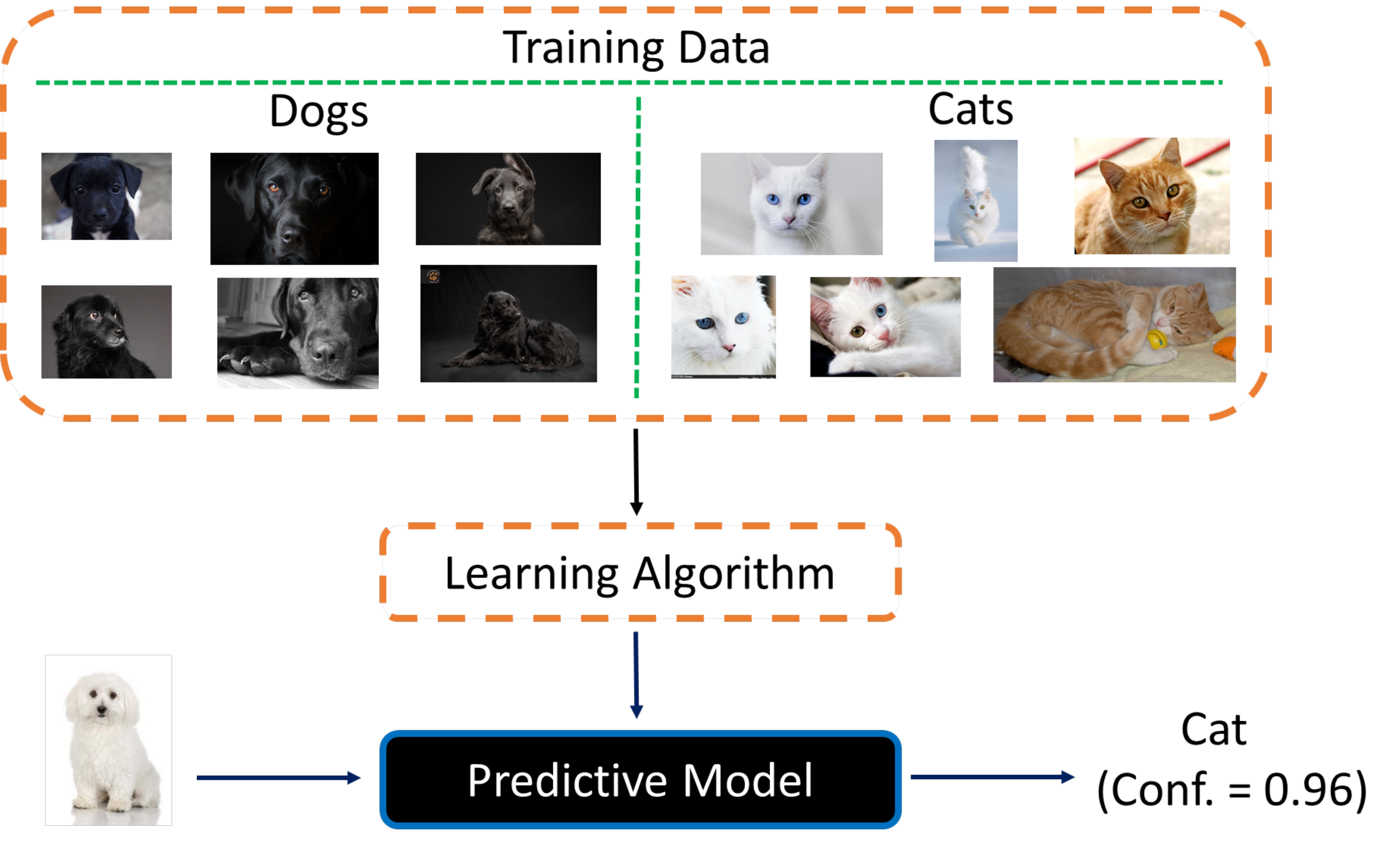}
\caption{Unknown unknowns in an image classification task. Training data comprised only of images of black dogs and of white and brown cats. A predictive model trained on this data incorrectly labels a \emph{white dog} (test image) as a \emph{cat} with high confidence.}
\label{fig:introfig}
\end{figure}

Unknown unknowns can arise when data that is used for training a predictive model is not representative of the samples encountered at test time when the model is deployed. This mismatch could be a result of unmodeled biases in the collection of training data or differences between the train and test distributions due to temporal, spatial or other factors such as a subtle shift in task definition. To illustrate, consider an image classification task where the goal is to predict if a given image corresponds to a cat or a dog (Figure \ref{fig:introfig}). Let us assume that the training data is comprised of images of black dogs, and white and brown cats, and the feature set includes details such as nose shape, presence or absence of whiskers, color, and shape of the eyes. A predictive model trained on such data might learn to make predictions solely based on color despite the presence of other discriminative features because color can perfectly separate the two classes in the training data. However, during test time, such a model would classify an image of a white dog as a cat with high confidence. The images of white dogs are, therefore, unknown unknowns with regard to such a predictive model. 

We formulate and address the problem of informed discovery of unknown unknowns of any given predictive model when deployed \emph{in the wild}. More specifically, we seek to identify unknown unknowns which occur as a result of systematic biases in the training data. We formulate this as an optimization problem where unknown unknowns are discovered by querying an oracle for true labels of selected instances under a fixed budget which limits the number of queries to the oracle. The formulation assumes no knowledge of the functional form or the associated training data of the predictive model and treats it as a black box which outputs a label and a confidence score (or a proxy) for a given data point. These choices are motivated by real-world scenarios in domains such as healthcare and judiciary, where predictive models are being deployed in settings where end users have no access to either the model details or the associated training data (e.g., COMPAS risk assessment tool for sentencing~\cite{brennan2009evaluating}). Identifying the blind spots of predictive models in such high-stakes settings is critical as undetected unknown unknowns can be catastrophic. In criminal justice, biases and blindspots can lead to the inappropriate sentencing or incarceration of people charged with crimes or unintentional racial biases~\cite{katearticle}. To the best of our knowledge, this is the first work providing an algorithmic approach to addressing this problem.

Developing an algorithmic solution for the discovery of unknown unknowns introduces a number of challenges:
1) Since unknown unknowns can occur in any portion of the feature space, how do we develop strategies which can effectively and efficiently search the space? 2) As confidence scores associated with model predictions are typically not informative for identifying unknown unknowns, how can we make use of the feedback from an oracle to guide the discovery of unknown unknowns? 3) How can we effectively manage the trade-off between searching in neighborhoods where we previously found unknown unknowns and examining unexplored regions of the search space? 

%a summary of the approach
To address the problem at hand, we propose a two-step approach which first partitions the test data such that instances with similar feature values and confidence scores assigned by the predictive model are grouped together, and then employs an explore-exploit strategy for discovering unknown unknowns across these \partitions based on the feedback from an oracle. The first step, which we refer to as \emph{\DSP} (\DSPshort), is guided by an objective function which encourages partitioning of the search space such that instances within each \partition are maximally similar in terms of their feature values and confidence scores. \DSPshort also provides interpretable explanations of the generated \partitions by associating a comprehensible and compact description with each \partition. As we later demonstrate in our experimental results, these interpretable explanations are very useful in understanding the properties of unknown unknowns discovered by our framework. We show that our objective is NP-hard and outline a greedy solution which is a $ln$ $N$ approximation, where $N$ is the number of data points in the search space. 
The second step of our methodology facilitates an effective \emph{exploration} of the \partitions generated by \DSPshort while \emph{exploiting} the feedback from an oracle. We propose a multi-armed bandit algorithm, \emph{\UUB} (\UUBshort), which exploits problem-specific characteristics to efficiently discover unknown unknowns. 

% assumptions
%We build on the intuition that unknown unknowns, which may often occur due to systematic biases, are concentrated in certain portions of the feature space and do not typically occur at random. O
The proposed methodology builds on the intuition that unknown unknowns occurring due to systematic biases are often concentrated in certain specific portions of the feature space and do not occur randomly~\cite{btm}. For instance, the example in Figure~\ref{fig:introfig} illustrates a scenario where systematic biases in the training data caused the predictive model to wrongly infer color as the distinguishing feature. Consequently, images following a specific pattern (i.e., all of the images of white dogs) turn out to be unknown unknowns for the predictive model. Another key assumption that is crucial to the design of effective algorithmic solutions for the discovery of unknown unknowns is that available evidential features are informative enough to characterize different subsets of unknown unknowns.
If such features were not available in the data, it would not be possible to leverage the properties of previously discovered unknown unknowns to find new ones. Consequently, learning algorithms designed to discover unknown unknowns would not be able to perform any better than blind search (no free lunch theorem~\cite{wolpert1997no}).

%empirical evaluation
We empirically evaluate the proposed framework on the task of discovering unknown unknowns occurring due to a variety of factors such as biased training data and domain adaptation across various diverse tasks, such as sentiment classification, subjectivity detection, and image classification. We experiment with a variety of base predictive models, ranging from decision trees to neural networks.
The results demonstrate the effectiveness of the framework and its constituent components for the discovery of unknown unknowns across different experimental conditions, providing evidence that the method can be readily applied to discover unknown unknowns in different real-world settings. 

\section{Problem Formulation}\label{probdef}
Given a black-box predictive model $\mathcal{M}$ which takes as input a data point $x$ with features $\mathcal{F} = \{f_1, f_2, \cdots f_L\}$, and returns a class label $c' \in C$ and a confidence score $s \in \left[0, 1\right]$, our goal is to find the unknown unknowns of $\mathcal{M}$ w.r.t a given test set $\mathcal{D}$ using a limited number of queries, $B$, to the oracle, and, more broadly, to maximize the utility associated with the discovered unknown unknowns. The discovery process is guided by a utility function, which not only incentivizes the discovery of unknowns unknowns, but also accounts for the costs associated with querying the oracle (e.g., monetary and time costs of labeling in crowdsourcing). Recall that, in this work, we focus on identifying unknown unknowns arising due to systematic biases in the training data. It is important to note that our formulation not only treats the predictive model as a black-box but also assumes no knowledge about the data used to train the predictive model.

Although our methodology is generic enough to find unknown unknowns associated with all the classes in the data, we formulate the problem for a particular class $c$, a \emph{critical class}, where false positives are costly and need to be discovered~\cite{elkan2001foundations}.  Based on the decisions of the system designer regarding critical class $c$ and confidence threshold $\tau$, our search space for unknown unknown discovery constitutes all of those data points in $\mathcal{D}$ which are assigned the critical class $c$ by model $\mathcal{M}$ with confidence higher than $\tau$ .

Our approach takes the following inputs: 1) A set of $N$ instances, $\mathcal{X} = \{x_1, x_2 \cdots x_N \} \subseteq \mathcal{D}$, which were \emph{confidently} assigned to the critical class $c$ by the model $\mathcal{M}$, and the corresponding confidence scores, $\mathcal{S} = \{s_1, s_2 \cdots s_N\}$, assigned to these points by $\mathcal{M}$, %and the confidence threshold $\tau$ which was used to filter these points. 
2) An oracle $o$ which takes as input a data point $x$ and returns its true label $o(x)$ as well as the cost incurred to determine the true label of $x$, $cost(x)$ 3) A budget $B$ on the number of times the oracle can be queried. 

%We consider each query to the oracle as a \emph{time step} which implies that the budget $B$ ensures that there are $B$ time steps in the entire process. 
Our utility function, $u(x(t))$, for querying the label of data point $x(t)$ at the $t^{th}$ step of exploration is defined as:
\begin{equation}\label{eqn:uidefn}
u(x(t)) = \mathds{1}_{\{o(x_t) \neq c\}} - \gamma \times cost(x(t))
\end{equation}
where $\mathds{1}_{\{o(x_t) \neq c\}}$ is an indicator function which returns $1$ if $x(t)$ is identified as an unknown unknown, and a $0$ otherwise. $cost(x(t))\in \left[0, 1\right]$ is the cost incurred by the oracle to determine the label of $x(t)$. 
Both the indicator and the cost functions in Equation \ref{eqn:uidefn} are initially unknown and observed based on oracle's feedback on $x(t)$.  $\gamma \in \left[0, 1\right]$ is a tradeoff parameter which can be provided by the end user. 
\\

\textbf{Problem Statement:}  
Find a sequence of $B$ instances $\{ x(1), x(2) \cdots x(B)\} \subseteq \mathcal{X}$ for which the cumulative utility $\sum\limits_{t=1}^{B} u(x(t))$ is maximum.

\section{Methodology}
In this section, we present our two-step framework designed to address the problem of informed discovery of unknown unknowns which occur due to systematic biases in the training data. We begin this section by highlighting the assumptions required for our algorithmic solution to be effective:
\begin{enumerate}
\item Unknown unknowns arising due to biases in training data typically occur in certain specific portions of the feature space and not at random. For instance, in our image classification example, the systematic bias of not including white dog images in the training data resulted in a specific category of unknown unknowns which were all clumped together in the feature space and followed a specific pattern. Attenberg et. al.~\cite{btm} observed this assumption to hold in practice and leveraged human intuition to find systematic patterns of unknown unknowns. 
\item We also assume that the features available in the data can effectively characterize different kinds of unknown unknowns, but the biases in the training data prevented the predictive model from leveraging these discriminating features for prediction. If such features were not available in the data, it would not be possible to utilize the characteristics of previously discovered unknown unknowns to find new ones. Consequently, no learning algorithm would perform better than blind search if this assumption did not hold (no free lunch theorem~\cite{wolpert1997no}).  
\end{enumerate}

Below we discuss our methodology in detail. First we present \emph{\DSP} (\DSPshort), which induces a similarity preserving partition on the set $\mathcal{X}$. Then, we present a novel multi-armed bandit algorithm, which we refer to as \emph{\UUB} (\UUBshort), for systematically \emph{searching} for unknown unknowns across these partitions while leveraging feedback from an oracle. 
\subsection{Descriptive Space Partitioning}
Our approach exploits the aforementioned intuition that blind spots arising due to systematic biases in the data do not occur at random, but are instead concentrated in specific portions of the feature space. 
The first step of our approach, \DSPshort, partitions the instances in $\mathcal{X}$ such that instances which are grouped together are similar to each other w.r.t the feature space $\mathcal{F}$ and were assigned similar confidence scores by the model $\mathcal{M}$. Partitioning $\mathcal{X}$ enables our bandit algorithm, \UUBshort, to discover regions with high concentrations of unknown unknowns. 
\begin{algorithm}
\small
\caption{Greedy Algorithm for Partitioning}
\label{alg:partitioning}
\begin{algorithmic}[1]
\State \textbf{Input:} Set of instances $\mathcal{X}$, Confidence scores $\mathcal{S}$, Patterns $\mathcal{Q}$, Metric functions $\{g_1 \cdots g_5\}$, Weights $\lambda$
\State \textbf{Procedure:}
\State $\mathcal{P} = \emptyset$, $\mathcal{E} = \mathcal{X}$
\While{$\mathcal{E} \neq \emptyset$}:
\State   \[p = \argmax\limits_{q \in \mathcal{Q}} \frac{|\mathcal{E} \cap  covered\_by(q)|}{g(q)}\]
where \[ g(q) = \lambda_1 g_1(q) - \lambda_2 g_2(q) + \lambda_3 g_3(q) - \lambda_4 g_4(q) + \lambda_5 g_5(q)\]
\State $\mathcal{P} = \mathcal{P} \cup p$ \text{, } $\mathcal{Q} = \mathcal{Q}~\backslash~p$
\text{, } $\mathcal{E} = \mathcal{E}~\backslash~covered\_by(p)$
\EndWhile
\State \Return $\mathcal{P}$
\end{algorithmic}
\end{algorithm}
\normalsize 

The intuition behind our partitioning approach is that two instances $a$ and $a'\in \mathcal{X}$ are likely to be judged using a similar logic by model $\mathcal{M}$ if they share similar feature values and are assigned to the same class $c$ with comparable confidence scores by $\mathcal{M}$. In such cases, if $a$ is identified as an unknown unknown, $a'$ is likely to be an unknown unknown as well\footnote{Note that this is not always the case, as we will see in the next section.}.
%since the model $\mathcal{M}$ employs similar logic in assigning these points to class $c$. 
Based on this intuition, we propose an objective function which encourages grouping of instances in $\mathcal{X}$ that are \emph{similar} w.r.t the criteria outlined above, and facilitates separation of \emph{dissimilar} instances. The proposed objective also associates a concise, comprehensible description with each partition, which is useful for understanding the exploration behavior of our framework and the kinds of unknown unknowns of $\mathcal{M}$ (details in the Experimental Evaluation Section).

\DSPshort takes as input a set of candidate patterns $\mathcal{Q} = \{q_1, q_2, \cdots\}$ where each $q_i$ is a conjunction of (feature, operator, value) tuples where operator $\in \{=, \neq, \leq, <, \geq, >\}$. Such patterns can be obtained by running an off-the-shelf frequent pattern mining algorithm such as Apriori~\cite{agrawal1994fast} on $\mathcal{X}$. Each pattern \emph{covers} a set of one or more instances in $\mathcal{X}$. For each pattern $q$, the set of instances that satisfy $q$ is represented by $covered\_by(q)$,  the centroid of such instances is denoted by $\bar{x}_q$, and their mean confidence score is $\bar{s}_q$.

The partitioning objective minimizes dissimilarities of instances within each \partition and maximizes them across \partitions. In particular, we define \emph{goodness} of each pattern $q$ in $\mathcal{Q}$ as the combination of following metrics, where $d$ and $d'$ are standard distance measures defined over feature vectors of instances and their confidence scores respectively:
\begin{flalign*}
& \text{\textbf{Intra-partition feature distance:}} & \\ 
& g_1(q) = \sum\limits_{\{x \in \mathcal{X} : \text{ } x \text{ } \in \text{ } covered\_by(q)\}} d(x, \bar{x}_q) & \\
& \text{\textbf{Inter-partition feature distance:}} & \\ 
& g_2(q) = \sum\limits_{\{x \in \mathcal{X} : \text{ } x \text{ } \in \text{ } covered\_by(q)\}} \sum\limits_{ \{q' \in \mathcal{Q}: \text{ } q' \neq q\}} d(x, \bar{x}_{q'}) & \\
& \text{\textbf{Intra-partition confidence score distance:}} & \\ 
& g_3(q) = \sum\limits_{\{s_i: \text{ } x_i \in \mathcal{X} \text{ } \wedge x_i \text{ } \in \text{ } covered\_by(q)\}} d'(s_i, \bar{s}_q) & \\
& \text{\textbf{Inter-partition confidence score distance:}} & \\ 
& g_4(q) = \sum_{\substack{\{s_i: \text{ } x_i \in \mathcal{X} \text{ } \wedge \\ x_i \text{ } \in \text{ } covered\_by(q)\}}} \sum\limits_{ \{q' \in \mathcal{Q}: \text{ } q' \neq q\}} d'(s_i, \bar{s}_{q'}) & \\
& \text{\textbf{Pattern Length: }} g_5(q) = size(q) \text{, the number of }&
\\ &\text{ (feature, operator, value) tuples in pattern $q$, included to } &
\\ &\text{ favor concise descriptions.}
\end{flalign*}
Given the sets of instances $\mathcal{X}$, corresponding confidence scores $\mathcal{S}$, a collection of patterns $\mathcal{Q}$, and weight vector $\lambda$ used to combine $g_1$ through $g_5$, our goal is to find a set of patterns $\mathcal{P} \subseteq \mathcal{Q}$ such that it covers all the points in $\mathcal{X}$ and minimizes the following objective:
%do not remove the label of the equation below, we are using it later in expts.

\footnotesize
\begin{align}\label{eqn:partobj}
min \sum\limits_{q \in Q} f_q (\lambda_1 g_1(q)-\lambda_2 g_2(q) + \lambda_3 g_3(q) \nonumber \\
- \lambda_4 g_4(q) + \lambda_5 g_5(q)) \\
\text{s.t.}\sum\limits_{q:\text{ }x \in covered\_by(q)} f_q \geq 1 \text{  } \forall x \in \mathcal{X}\text{, where }
f_q \in \{0, 1\} \text{  } \nonumber \\
\forall q \in \mathcal{Q} \nonumber
\end{align}
\normalsize 
where $f_q$ corresponds to an indicator variable associated with pattern $q$ which determines if the pattern $q$ has been added to the solution set ($f_q = 1$) or not ($f_q = 0$). 

The aforementioned formulation is identical to that of a weighted set cover problem which is NP-hard~\cite{johnson1974approximation}. It has been shown that a greedy solution provides a $ln$ $N$ approximation to the weighted set cover problem~\cite{johnson1974approximation,feige1998threshold} where $N$ is the size of search space. 
Algorithm \ref{alg:partitioning} applies a similar strategy which greedily selects patterns with maximum coverage-to-weight ratio at each step, thus resulting in a $ln$ $N$ approximation guarantee. This process is repeated until no instance in $\mathcal{X}$ is left uncovered. 
If an instance in $\mathcal{X}$ is covered by multiple \partitions, ties are broken by assigning it to a \partition with the closest centroid.

Our partitioning approach is inspired by a class of clustering techniques commonly referred to as conceptual clustering~\cite{michalski1983learning,fisher1987knowledge} or descriptive clustering~\cite{weiss2006descriptive,li2008new,kim2014bayesian,lakkaraju2016confusions}. 
We make the following contributions to this line of research: We propose a novel objective function, whose components have not been jointly considered before. In contrast to previous solutions which employ post processing techniques or use Bayesian frameworks, we propose a simple, yet elegant solution which offers theoretical guarantees. 

\subsection{Multi-armed Bandit for Unknown Unknowns}
The output of the first step of our approach, DSP, is a set of $K$ \partitions $\mathcal{P} = \{p_1, p_2 \cdots p_K\}$ such that each $p_j$ corresponds to a set of data points which are similar w.r.t the feature space $\mathcal{F}$ and have been assigned similar confidence scores by the model $\mathcal{M}$. 
The partitioning scheme, however, does not guarantee that all data points in a \partition share the same characteristic of being unknown unknown (or not being unknown unknown). 
It is important to note that sharing similar feature values and confidence scores does not ensure that the data points in a \partition are indistinguishable as far as the model logic is concerned. This is due to the fact that the model $\mathcal{M}$ is a black-box and we do not actually observe the underlying functional forms and/or feature importance weights being used by $\mathcal{M}$. Consequently, each \partition has an unobservable concentration of unknown unknown instances. 
The goal of the second step of our approach is to compute an exploration policy over the \partitions generated by \DSPshort such that it maximizes the cumulative utility of the discovery of unknown unknowns (as defined in the Problem Formulation section).

We formalize this problem as a multi-armed bandit problem and propose an algorithm for deciding which \partition to query at each step (See Algorithm 2). In this formalization, each \partition $p_j$ corresponds to an arm $j$ of the bandit. At each step, the algorithm chooses a \partition and then randomly samples a data point from that \partition without replacement and queries its true label from the oracle. Since querying the data point reveals whether it is an unknown unknown, the point is excluded from future steps. 

\begin{algorithm}[t]
\small
\caption{Explore-Exploit Algorithm for Unknown Unknowns}
\label{alg:uub}
\begin{algorithmic}[1]
\State \textbf{Input:}
\State Set of \partitions (arms) $\{1, 2 \cdots K\}$, Oracle $o$, Budget $B$
\State \textbf{Procedure:}
\For {t from 1 to B}:
\If {$t \leq K$}:
\State Choose arm $A_t = t$
\Else
\State Choose arm $A_t = \argmax\limits_{1 \leq i \leq K} \bar{u}_t(i) + b_{t}(i)$ 
\EndIf
\State Sample an instance $x(t)$ from \partition $p_{A_t}$ and query the oracle for its true label
\State Observe true label of $x(t)$ and the cost of querying the oracle and compute $u(x(t))$ using Equation (\ref{eqn:uidefn}).
\EndFor
\State \Return $\sum\limits_{t=1}^{B} u(x(t))$
\end{algorithmic}
\end{algorithm}
\normalsize
In the first $K$ steps, the algorithm samples a point from each \partition. Then, at each step $t$, the exploration decisions are guided by a combination of $\bar{u}_t(i)$, the empirical mean utility (reward) of the \partition $i$ at time $t$, and $b_{t}(i)$, which represents the uncertainty over the estimate of $\bar{u}_t(i)$. 

Our problem setting has the characteristic that the expected utility of each arm is non-stationary; querying a data point from a \partition changes the concentration of unknown unknowns in the \partition and consequently changes the expected utility of that \partition in future steps. Therefore, stationary MAB algorithms such as UCB~\cite{auer2002finite} are not suitable. 
A variant of UCB, \emph{discounted UCB}, addresses the non-stationary settings and can be used as follows to compute $\bar{u}_t(i)$ and $b_{t}(i)$ \cite{garivier2008upper}. 
\begin{align*}\label{eqn:UCB}
\bar{u}_t(i) = \frac{1}{N_t(\vartheta_{t}^i,i)} \sum\limits_{j=1}^{t} \vartheta_{j,t}^i \text{ }u(x(j))\text{ }\mathds{1}_{A_j = i} \\ 
b_t(i) = \sqrt{\frac{2 \text{ }log \sum\limits_{i=1}^{K} N_t(\vartheta_{t}^i, i)}{N_t(\vartheta_{t}^i,i)}} \text{, } N_t(\vartheta_{t}^i,i) = \sum\limits_{j=1}^{t} \vartheta_{j,t}^i \text{ }\mathds{1}_{A_j = i}
\end{align*}
The main idea of discounted UCB is to weight recent observations more to account for the non-stationary nature of the utility function. If $\vartheta_{j,t}^i$ denotes the discounting factor applied at time $t$ to the reward obtained from arm $i$ at time $j < t$, $\vartheta_{j,t}^i = \gamma^{t-j}$ in the case of discounted UCB, where $\gamma \in (0,1)$. Garivier et. al. established a lower bound on the regret in the presence of abrupt changes in the reward distributions of the arms and also showed that discounted UCB matches this lower bound upto a logarithmic factor~\cite{garivier2008upper}. 

The discounting factor of discounted UCB is designed to handle arbitrary changes in the utility distribution, whereas the way the utility of a \partition changes in our setting has a certain structure: The utility estimate of arm $i$ only changes by a bounded quantity when the arm is queried.
Using this observation, we can customize the calculation of $\vartheta_{j,t}^i$ for our setting and eliminate the need to set up the value of $\gamma$, which affects the quality of decisions made by discounted UCB. 
We compute $\vartheta_{j,t}^i$ as the ratio of the number of data points in the partition $i$ at time $j$ to the number of data points in the partition $i$ at time $t$:
\begin{equation}\label{eqn:eqscore}
\vartheta_{j,t}^i = (\mathcal{N}_i - \sum\limits_{l=1}^{t} \mathds{1}_{A_l = i}) \text{ } \big{/} \text{ } (\mathcal{N}_i - \sum\limits_{l=1}^{j} \mathds{1}_{A_l = i}) 
\end{equation}
The value of $\vartheta_{j,t}^i$ is inversely proportional to the number of pulls of arm $i$ during the interval $(j, t)$. $\vartheta_{j,t}^i$ is 1, if the arm $i$ is not pulled during this interval, indicating that the expected utility of $i$ remained unchanged. 
We refer to the version of Algorithm 2 that uses the discounting factor specific to our setting (Eqn. \ref{eqn:eqscore}) as Bandit for Unknown Unknowns (\UUBshort).

\section{Experimental Evaluation}\label{experiments}
We now present details of the experimental evaluation of constituent components of our framework as well as the entire pipeline. 
\\
\textbf{Datasets and Nature of Biases:}
We evaluate the performance of our methodology across four different data sets in which the underlying cause of unknown unknowns vary from biases in training data to domain adaptation:\\
(1) \textit{Sentiment Snippets:} A collection of 10K sentiment snippets/sentences expressing opinions on various movies~\cite{Pang+Lee:05a}. Each snippet (sentence) corresponds to a data point and is labeled as positive or negative. 
We split the data equally into train and test sets. We then bias the training data by randomly removing \emph{sub-groups} of negative snippets from it. We consider \emph{positive sentiment} as the critical class for this data. \\ 
(2) \textit{Subjectivity:} A set of 10K subjective and objective snippets extracted from Rotten Tomatoes webpages~\cite{Pang+Lee:04a}. We consider the \emph{objective} class in this dataset as the critical class, split the data equally into train and test sets, and introduce bias in the same way as described above. \\
(3) \textit{Amazon Reviews:} A random sample of 50K reviews of books and electronics collected from Amazon~\cite{mcauley2015inferring}. We use this data set to study unknown unknowns introduced by domain adaptation; we train the predictive models on the electronics reviews and then \emph{test} them on the book reviews. Similar to the \emph{sentiment snippets} data set, the \emph{positive sentiment} is the critical class.\\ 
(4) \textit{Image Data:} A set of 25K cat and dog images~\cite{imagedata}. We use this data set to assess whether our framework can recognize unknown unknowns that occur when semantically meaningful sub-groups are missing from the training data. To this end, we split the data equally into train and test and bias the training data such that it comprises only of images of dogs which are black, and cats which are not black. We set the class label \emph{cat} to be the critical class in our experiments. \\
\textbf{Experimental Setting:}
We use bag of words features to train the predictive models for all of our textual data sets. As the features for the images, we use super-pixels obtained using the standard algorithms~\cite{ribeiro2016should}. Images are represented with a feature vector comprising of 1's and 0's indicating the presence or absence of the corresponding super pixels. We experiment with multiple predictive models: decision trees, SVMs, logistic regression, random forests and neural network. Due to space constraints, this section  presents results for decision trees as model $\mathcal{M}$ but detailed results for all the other models are included in the Appendix. We set the threshold for confidence scores $\tau$ to 0.65 to construct our search space $\mathcal{X}$ for each data set. We consider two settings for the cost function (refer Eqn. \ref{eqn:uidefn}): The cost is set to $1$ for all instances (uniform cost) in the image dataset and it is set to $\left[(\text{length}(x) - \text{minlength})/(\text{maxlength} - \text{minlength})\right]$ (variable cost) for all textual data. length$(x)$ denotes the number of words in a snippet (or review) $x$; minlength and maxlength denote the minimum and maximum number of words in any given snippet (or review). Note that these cost functions are only available to the oracle. The tradeoff parameter $\gamma$ is set to 0.2. The parameters of \DSPshort  $\{ \lambda_1, \cdots \lambda_5\}$ are estimated by setting aside as a validation set 5\% of the test instances assigned to the critical class by the predictive models. We search the parameter space using coordinate descent to find parameters which result in the minimum value of the objective function defined in Eqn. \ref{eqn:partobj}. We set the budget $B$ to $20\%$ of all the instances in the set $\mathcal{X}$ through out our experiments. Further, the results presented for \UUBshort are all averaged across $100$ runs.
%\begin{comment}
\subsection{Evaluating the Partitioning Scheme}
\begin{figure}
\centering
\includegraphics[scale=0.16]{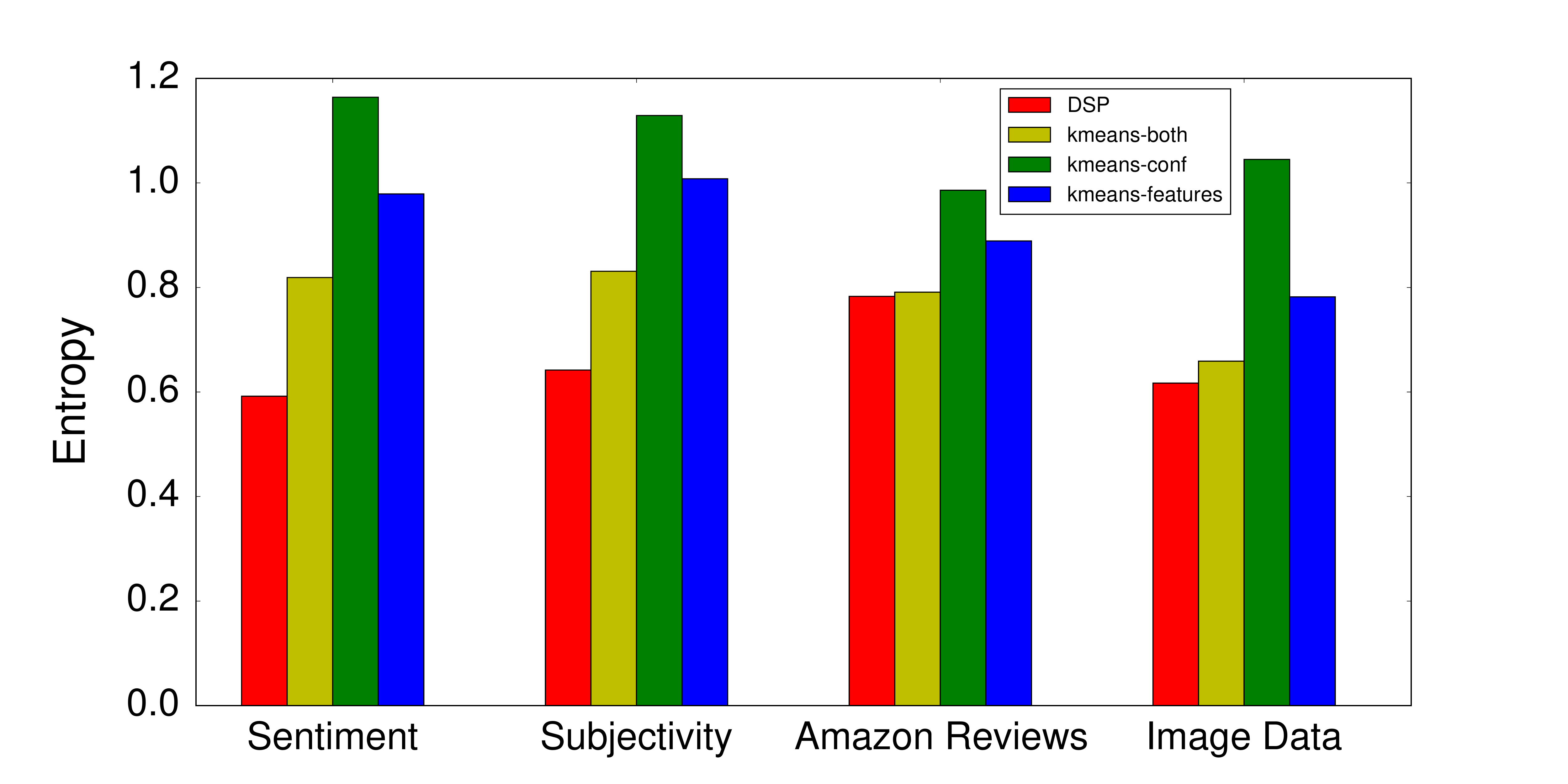}
\caption{Evaluating partitioning strategies using entropy (smaller values are better).}
\label{fig:entropy}
\end{figure}
%\end{comment}
The effectiveness of our framework relies on the notion that our partitioning scheme, \DSPshort, creates \partitions such that unknown unknowns are concentrated in a specific subset of \partitions as opposed to being evenly spread out across them. If unknown unknowns are distributed evenly across all the \partitions, our bandit algorithm cannot perform better than a strategy which randomly chooses a \partition at each step of the exploration process. We, therefore, measure the quality of \partitions created by \DSPshort by measuring the entropy of the distribution of unknown unknowns across the \partitions in $\mathcal{P}$. For each partition $p \in \mathcal{P}$, we count the number of unknown unknowns, $U_{p}$ based on the true labels which are only known to the oracle. We then compute entropy of $\mathcal{P}$ as follows: 
\[\text{Entropy}(\mathcal{P}) = - \sum\limits_{p \in \mathcal{P}} \frac{U_{p}}{\sum\limits_{p' \in \mathcal{P}} U_{p'}} log_2(\frac{U_{p}}{\sum\limits_{p' \in \mathcal{P}} U_{p'}} )\]
Smaller entropy values are desired as they indicate higher concentrations of unknown unknowns in fewer \partitions.
\begin{figure}[ht]
\centering
\subfigure[]{
    \includegraphics[scale=0.16]{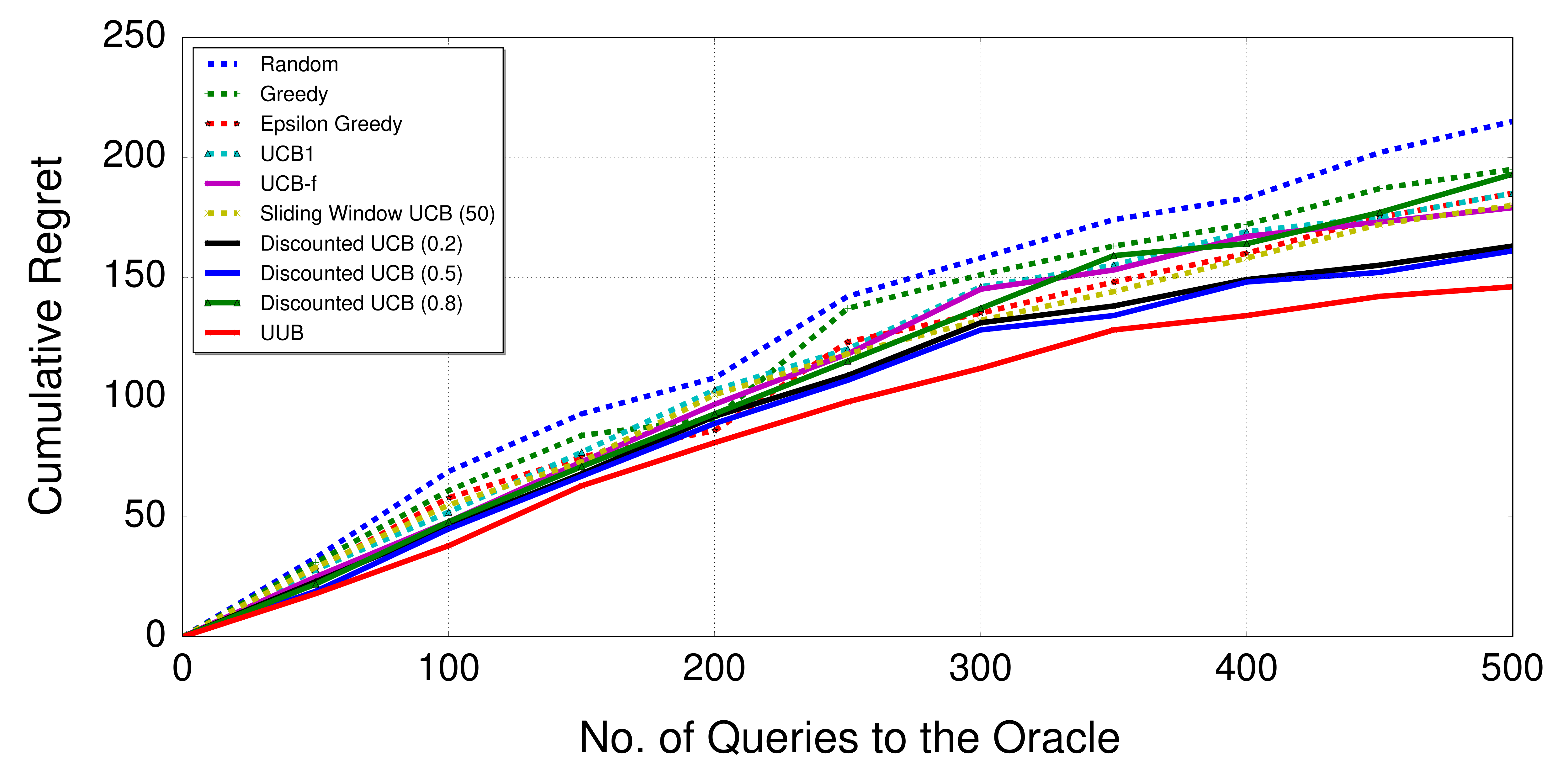}
    \label{fig:banditeval-image}
}
\subfigure[]{
    \includegraphics[scale=0.16]{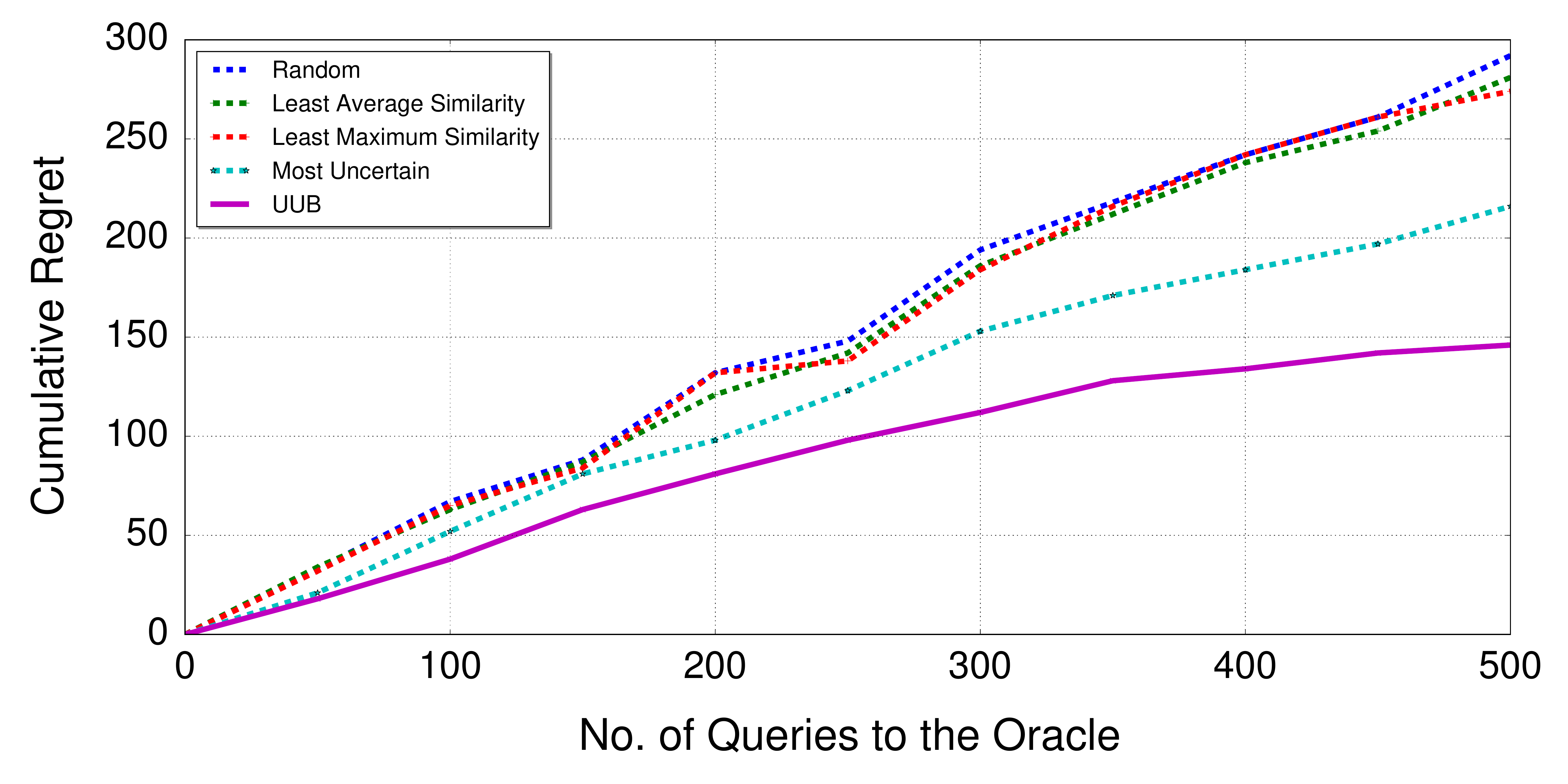}
    \label{fig:completeeval-image}
}
\caption[Optional caption for list of figures]{\subref{fig:banditeval-image} Evaluating the bandit framework on image data, \subref{fig:completeeval-image} Evaluating the complete pipeline on image data (decision trees as predictive model).}
\label{fig:combinedFigure}
\end{figure}

\begin{figure}
\centering
    \includegraphics[scale=0.18]{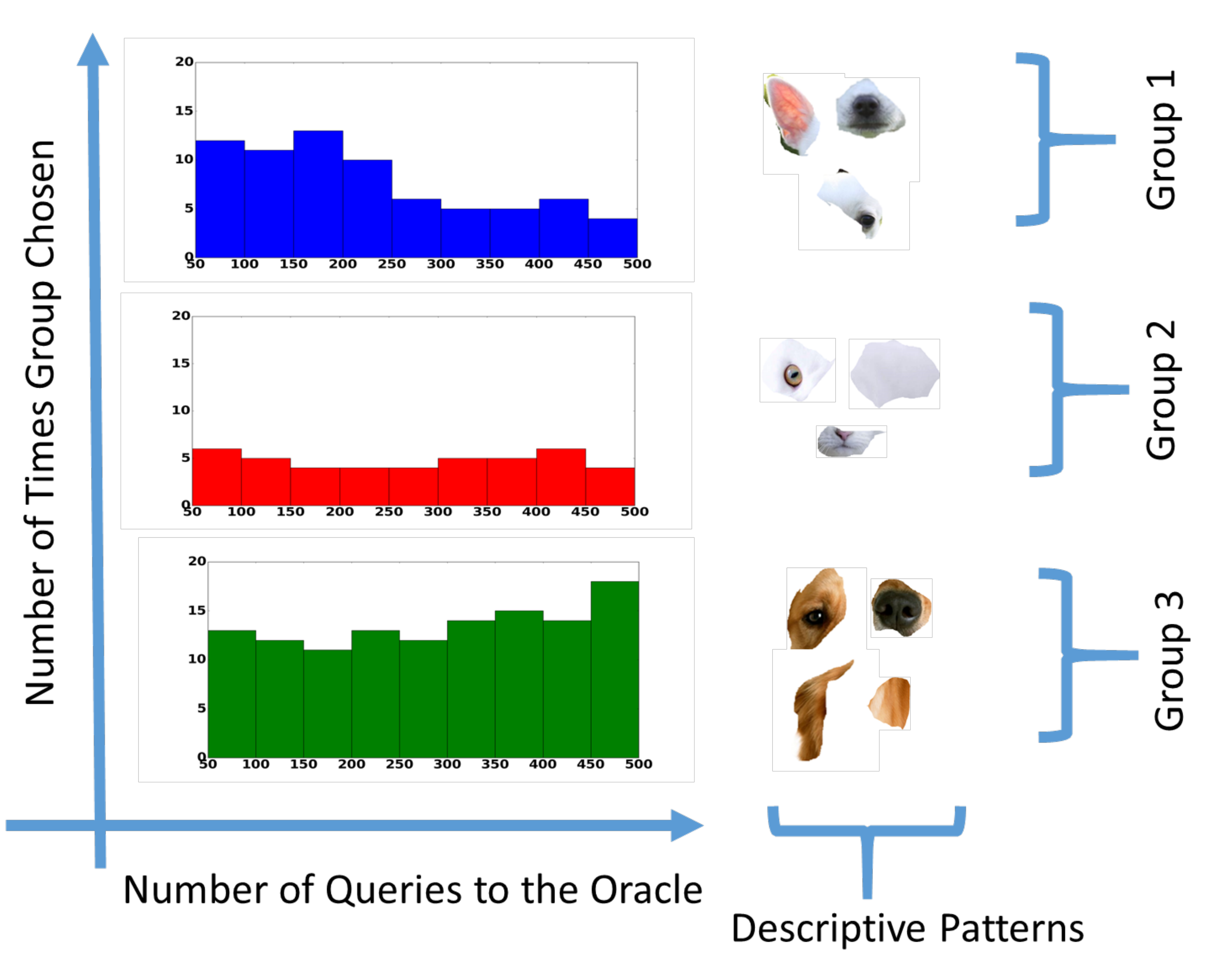}
\caption{Illustration of the methodology on image data.}
\label{fig:explore}
\end{figure}

%\ece{Now that we have the random baseline, it is confusing to call the other approaches (e.g., kmeans) as baselines as well. I made edits to the following paragraphs to address this issue.}
Figure \ref{fig:entropy} compares the entropy of the \partitions generated by \DSPshort with clusters generated by k-means algorithms using only features in $\mathcal{F}$ (kmeans-features), only confidence scores in $\mathcal{S}$ (kmeans-conf) and both (kmeans-both) by first clustering using confidence scores and then using features. The entropy values for \DSPshort are consistently smaller compared to alternative approaches using kmeans across all the datasets. This can be explained by the fact that \DSPshort jointly optimizes inter and intra-partition distances over both features and confidence scores. As shown in Figure \ref{fig:entropy}, the entropy values are much higher when k-means considers only features or only confidence scores indicating the importance of jointly reasoning about them. 

We also compare the entropy values obtained for \DSPshort as well as other k-means based approaches to an upper bound computed with random partitioning. For each of the algorithms (\DSPshort and other k-means based approaches), we designed a corresponding random partitioning scheme which randomly re-assigns all the data points in the set $\mathcal{X}$ to \partitions while keeping the number of \partitions and the number of data points within each \partition same as that of the corresponding algorithm. 
%The entropy associated with such a random partitioning scheme is likely to be quite high as unknown unknowns are randomly distributed across \partitions. 
We observe that the entropy values obtained for \DSPshort and all the other baselines are consistently smaller than those of the corresponding random partitioning schemes. Also, the entropy values for \DSPshort are about 32-37\% lower compared to its random counterpart across all of the datasets.

%We also computed the average inter-partition and intra-partition feature and confidence score  distances for \DSPshort on all the datasets and compared it to the baselines. We found that the inter-partition feature and confidence score distances for \DSPshort were on an average about 6.38\% and 4.12\% higher than the inter-partition feature and confidence score distances for the k-means clustering algorithm which uses both features and confidence scores. 
\subsection{Evaluating the Bandit Algorithm}
We measure the performance of our multi-armed bandit algorithm \UUBshort in terms of a standard evaluation metric in the MAB literature called \emph{cumulative regret}. Cumulative regret of a policy $\pi$ is computed as the difference between the total reward collected by an optimal policy $\pi^*$, which at each step plays the arm with the highest expected utility (or reward) and the total reward collected by the policy $\pi$. Small values of cumulative regret indicate better policies. The utility function defined in Eqn. \ref{eqn:uidefn} determines the reward associated with each instance. 

We compare the performance of our algorithm, \UUBshort, with that of several baseline algorithms such as random, greedy, $\epsilon$-greedy strategies~\cite{chapelle2011empirical}, UCB, UCB$_f$~\cite{slivkins2008adapting}, sliding window and discounted UCB~\cite{garivier2008upper} for various values of the discounting factor $\gamma =\{0.2, 0.5, 0.8\}$. 
All algorithms take as input the \partitions created by \DSPshort. 
Figure \ref{fig:combinedFigure}(a) shows the cumulative regret of each of these algorithms on the image data set. Results for the other data sets can be seen in the Appendix.  
The figure shows that \UUBshort achieves the smallest cumulative regret compared to other baselines on the image data set. 
Similarly, \UUBshort is the best performing algorithm on the sentiment snippets and subjectivity snippets data sets, whereas discounted UCB ($\gamma = 0.5$) achieves slightly smaller regret than \UUBshort on the Amazon reviews data set.
The experiment also highlights a disadvantage of the discounted UCB algorithm as its performance is sensitive to the choice of the discounting factor $\gamma$, where as \UUBshort is parameter free. Further, both UCB and its variant UCB$_f$ which are designed for stationary and slowly changing reward distributions respectively have higher cumulative regret than \UUBshort and discounted UCB indicating that they are not as effective in our setting. 

\subsection{Evaluating the Overall Methodology}
In the previous section, we compared the performance of \UUBshort to other bandit methods when they are given the same data \partitions to explore.  In this section, we evaluate the performance of our complete pipeline (\DSPshort + \UUBshort). Due to the lack of existing baselines which address the problem at hand, we compare the performance of our framework to other end-to-end heuristic methods we devised as baselines. Due to space constraints, we present results only for the image dataset. Results for other data sets can be seen in the Appendix.  

We compare the cumulative regret of our framework to that of a variety of baselines: 1) Random sampling: Randomly select $B$ instances from set $\mathcal{X}$ for querying the oracle. 2) Least average similarity: For each instance in $\mathcal{X}$, compute the average Euclidean distance w.r.t all the data points in the training set and choose $B$ instances with the largest distance. 3) Least maximum similarity: Compute minimum Euclidean distance of each instance in $\mathcal{X}$ from the training set and choose $B$ instances with the highest distances. 4) Most uncertain: Rank the instances in $\mathcal{X}$ in increasing order of the confidence scores assigned by the model $\mathcal{M}$ and pick the top $B$ instances. The least average similarity and least maximum similarity baselines are related to research on outlier detection~\cite{chandola2007outlier}. Furthermore, the baseline titled \emph{most uncertain} is similar to the uncertainty sampling query strategy used in active learning literature. Note that the least average similarity and the least maximum similarity baselines assume access to the data used to train the predictive model unlike our framework which makes no such assumptions. Figure \ref{fig:combinedFigure}(b) shows the cumulative regret of our framework and the baselines for the image data. It can be seen that \UUBshort achieves the least cumulative regret of all the strategies across all data sets. It is interesting to note that the least average similarity and the least maximum similarity approaches perform worse than \UUBshort in spite of having access to additional information in the form of training data. 
%It is interesting to note that the baseline leveraging the uncertainty of the model performs better than the baselines which exploit the dissimilarity of test instances w.r.t training data. 
\\

\noindent{\textbf{Qualitative Analysis}}
Figure \ref{fig:explore} presents an illustrative example of how our methodology explores three of the \partitions generated for the image data set. Our partitioning framework associated the super pixels shown in the Figure with each \partition. Examining the super pixels reveals that \partitions 1, 2 and 3 correspond to the images of white chihuahuas (dog), white cats, and brown dogs respectively. The plot shows the number of times the arms corresponding to these \partitions have been played by our bandit algorithm. The figure shows that \partition 2 is chosen fewer times compared to \partitions 1 and 3 --- because white cat images are part of the training data used by the predictive models and there are not many unknown unknowns in this \partition. On the other hand, white and brown dogs are not part of the training data and our bandit algorithm explores these \partitions often. Figure \ref{fig:explore} also indicates that \partition 1 was explored often during the initial plays but not later on. This is because there were fewer data points in that \partition and the algorithm had exhausted all of them after a certain number of plays. 

\section{Related Work}
In this section, we review prior research relevant to the discovery of unknown unknowns. 
\\

\noindent{{\bfseries Unknown Unknowns}} The problem of model incompleteness and the challenge of grappling with unknown unknowns in the real world has been coming to the fore as a critical topic in discussions about the utility of AI technologies~\cite{horvitz2008artificial}.
Attenberg et. al. introduced the idea of harnessing human input to identify unknown unknowns but their studies left the task of exploration and discovery completely to humans without any assistance~\cite{btm}. In contrast, we propose an algorithmic framework in which the role of the oracle is simpler and more realistic: The oracle is only queried for labels of selected instances chosen by our algorithmic framework. 
\\

\noindent{{\bfseries Dataset Shift}} 
A common cause of unknown unknowns is \emph{dataset shift}, which represents the mismatch between training and test distributions~\cite{Candela2009,jiang2007}. Multiple approaches have been proposed to address dataset shift, including importance weighting of training instances based on similarity to test set ~\cite{shimodaira2000improving}, online learning of prediction models~\cite{cesa2006prediction}, and learning models robust to adversarial actions ~\cite{teo2007convex,herbrich2004invariant,decoste2002training}. These approaches cannot be applied to our setting as they make  one or more of the following assumptions which limit their applicability to real-world settings: 1) the model is not a black box 2) the data used to train the predictive model is accessible 3) the model can be adaptively retrained. Further, the goal of this work is different as we study the problem of discovering unknown unknowns of models which are already deployed. 
\\

\noindent{{\bfseries Active Learning}} 
Active learning techniques aim to build highly accurate predictive models while requiring fewer labeled instances. These approaches typically involve querying an oracle for labels of certain selected instances and utilizing the obtained labels to adaptively retrain the predictive models~\cite{settles2010active}. Various query strategies have been proposed to choose the instances to be labeled (e.g.,  uncertainty sampling~\cite{lewis1994sequential,settles2010active}, query by committee~\cite{seung1992query}, expected model change~\cite{settles2008multiple}, expected error reduction~\cite{zhu2003combining}, expected variance reduction~\cite{zhang2000value}). Active learning frameworks were designed to be employed during the learning phase of a predictive model and are therefore not readily applicable to our setting where the goal is to find blind spots of a black box model which has already been deployed. Furthermore, query strategies employed in active learning are guided towards the discovery of \textit{known unknowns}, utilizing information from the predictive model to determine which instances should be labeled by the oracle. These approaches are not suitable for the discovery of unknown unknowns as the model is not aware of unknown unknowns and it lacks meaningful information towards their discovery.
\\

\noindent{{\bfseries Outlier Detection}} 
Outlier detection involves identifying individual data points (global outliers) or groups of data points (collective outliers) which either do not conform to a target distribution or are dissimilar compared to majority of the instances in the data~\cite{han2011data,chandola2007outlier}. Several parametric approaches~\cite{agarwal2007detecting,abraham1979bayesian,eskin2000anomaly} were proposed to address the problem of outlier detection. These methods made assumptions about the underlying data distribution, and characterized those points with a smaller likelihood of being generated from the assumed distribution, as outliers. Non-parametric approaches~\cite{eskin2000anomaly,eskingeometric,fawcett1997adaptive} which made fewer assumptions about the distribution of the data such as histogram based methods, distance and density based methods were also proposed to address this problem. Though unknown unknowns of any given predictive model can be regarded as collective outliers w.r.t the data used to train that model, the aforementioned approaches are not applicable to our setting as we assume no access to the training data.  

\section{Discussion \& Conclusions}
We presented an algorithmic approach to discovering unknown unknowns of predictive models. The approach assumes no knowledge of the functional form or the associated training data of the predictive models, thus, allowing the method to be used to build insights about the behavior of deployed predictive models. In order to guide the discovery of unknown unknowns, we partition the search space and then use bandit algorithms to identify \partitions with larger concentrations of unknown unknowns. To this end, we propose novel algorithms both for partitioning the search space as well as sifting through the generated \partitions to discover unknown unknowns.

We see several research directions ahead, including opportunities to employ alernate objective functions. For instance, the budget $\mathcal{B}$ could be defined in terms of the total cost of querying the oracle instead of the number of queries to the oracle.  Our method can also be extended to more sophisticated settings where the utility of some types of unknown unknowns decreases with time as sufficient examples of the type are discovered (e.g., after informing the engineering team about the discovered problem). In many settings, the oracle can be approximated via the acquisition of labels from crowdworkers, and the labeling noise of the crowd might be addressed by incorporating repeated labeling into our framework.

The discovery of unknown unknowns can help system designers when deploying predictive models in numerous ways. The partitioning scheme that we have explored provides interpretable descriptions of each of the generated \partitions. These descriptions could help a system designer to readily understand the characteristics of the discovered unknown unknowns and devise strategies to prevent errors or recover from them (e.g., silencing the model when a query falls into a particular \partition where unknown unknowns were discovered previously). Discovered unknown unknowns can further be used to retrain the predictive model which in turn can recognize its mistakes and even correct them. 

Formal machinery that can shine light on limitations of our models and systems will be critical in moving AI solutions into the open world--especially for high-stakes, safety critical applications. We hope that this work on an algorithmic approach to identifying unknown unknowns in predictive models will stimulate additional research on incompleteness in our models and systems. 

\section{Acknowledgments}
Himabindu Lakkaraju carried out this research during an internship at Microsoft Research. The authors would like to thank Lihong Li, Janardhan Kulkarni, and the anonymous reviewers for their insightful comments and feedback. 

\bibliographystyle{aaai}
\small
%\bibliography{unknowns}

\normalsize

\clearpage
\section{Appendix}
%\subsection{Additional Results}
% bandit - sentiment snippets
% complete framework 
\begin{figure}[ht]
\centering
\subfigure[]{
    \includegraphics[scale=0.10]{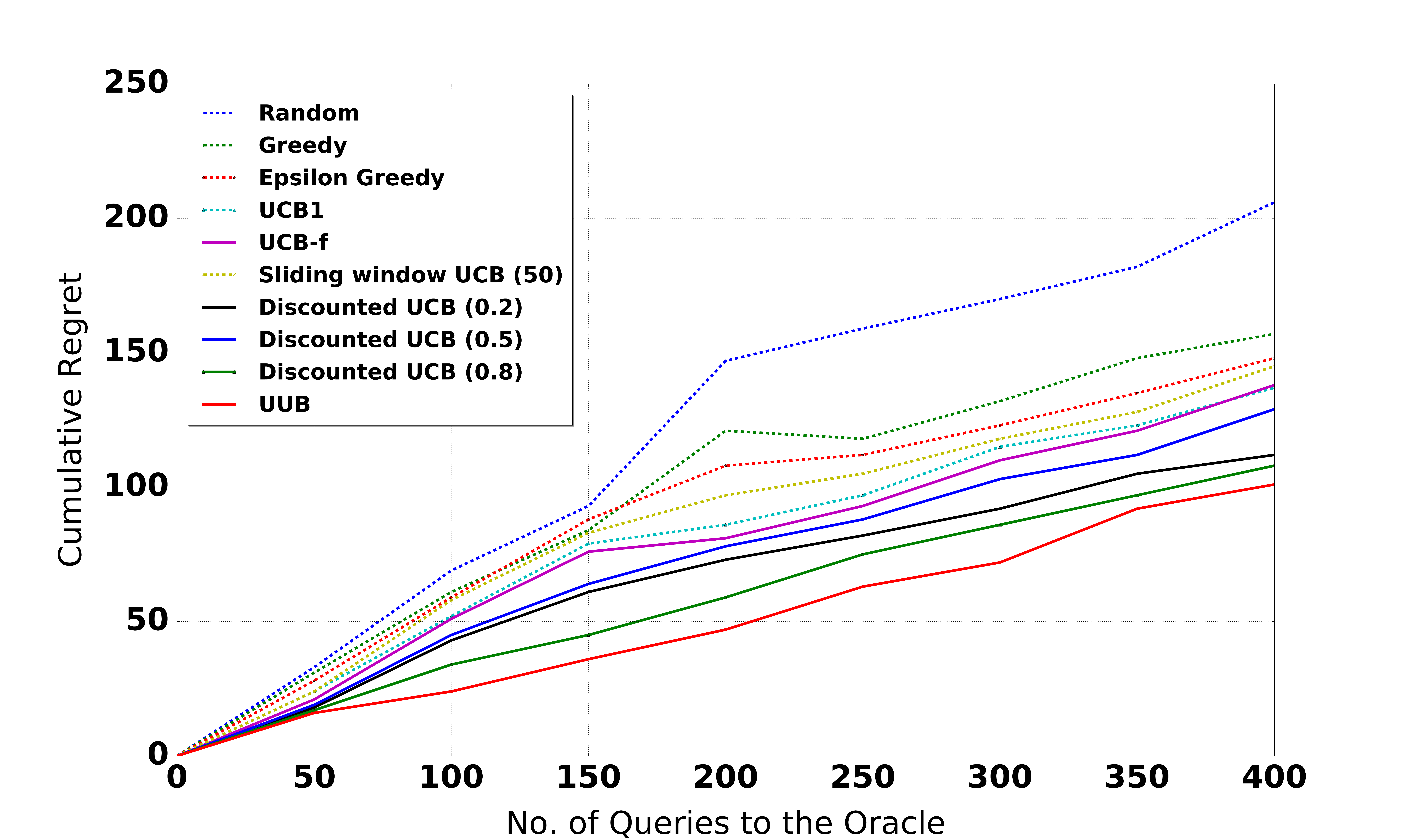}
    \label{fig:banditeval-senti}
}
\subfigure[]{
    \includegraphics[scale=0.10]{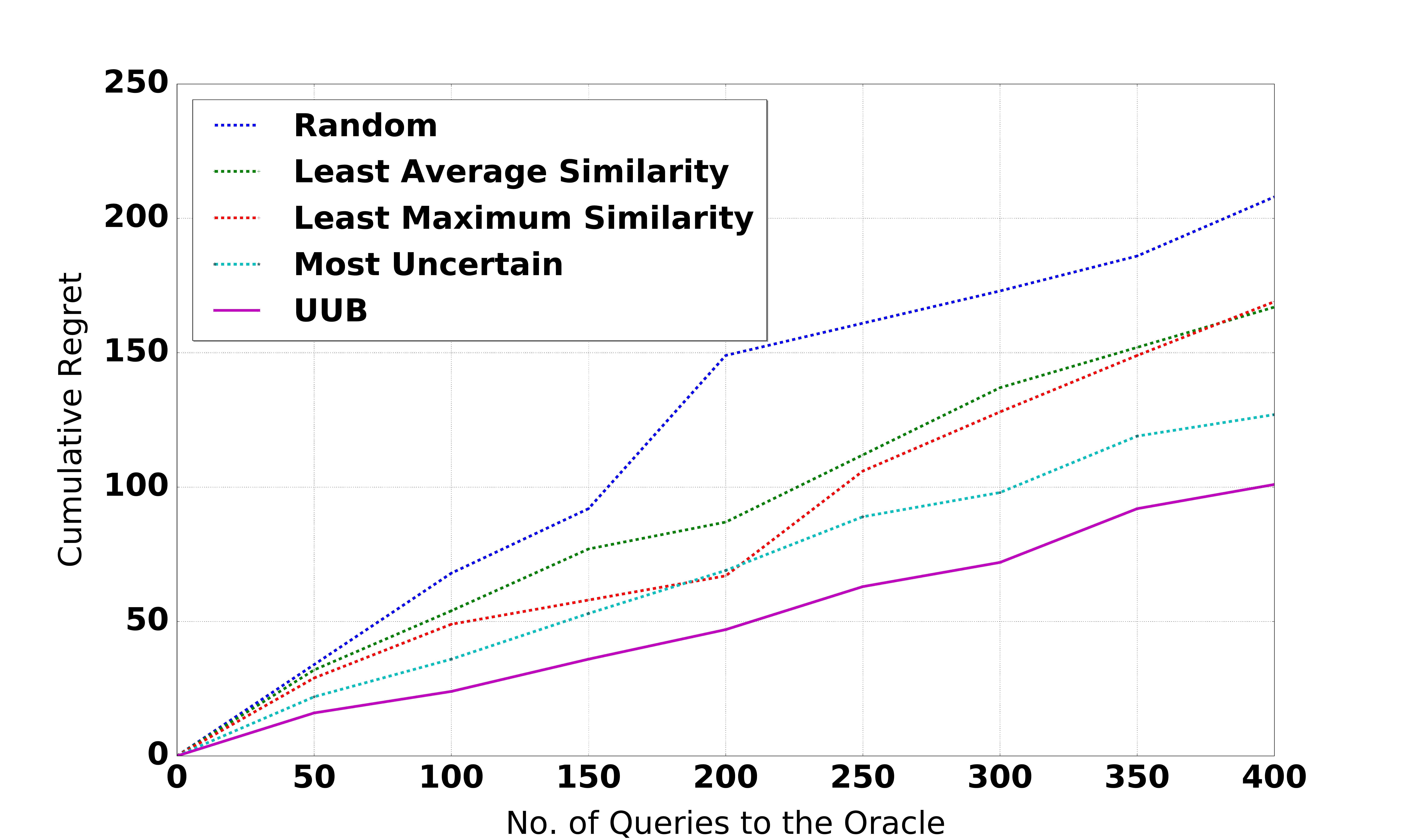}
    \label{fig:completeeval-senti}
}
\caption[Optional caption for list of figures]{\subref{fig:banditeval-senti} Evaluating the Bandit Framework on Sentiment Snippets, \subref{fig:completeeval-senti} Evaluating the Complete Pipeline on Sentiment Snippets [Decision trees as Predictive Model]}
\label{fig:subfigureExample}
\end{figure}
% bandit - subjectivity
% complete framework
\begin{figure}[ht]
\centering
\subfigure[]{
    \includegraphics[scale=0.10]{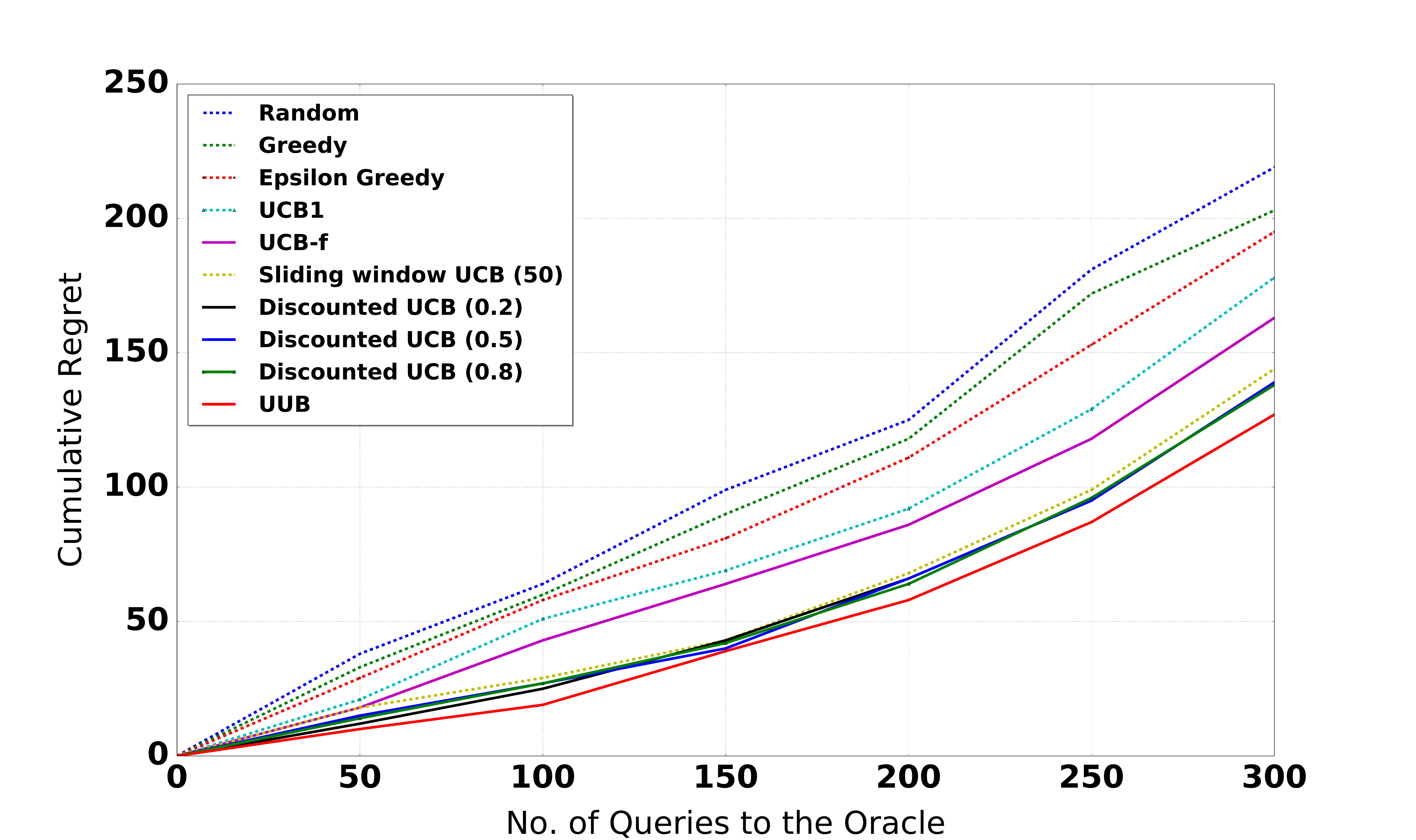}
    \label{fig:banditeval-subj}
}
\subfigure[]{
    \includegraphics[scale=0.10]{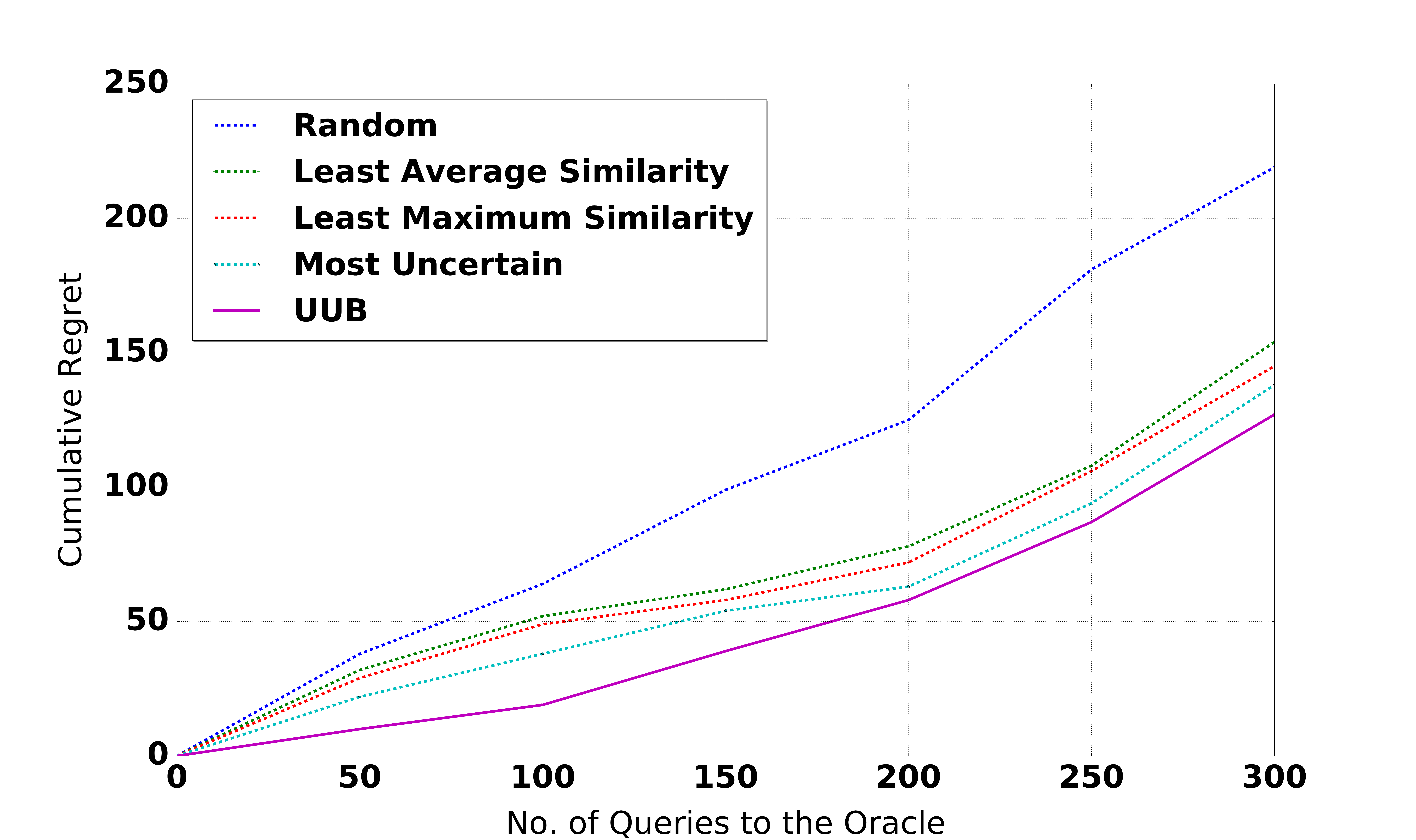}
    \label{fig:completeeval-subj}
}
\caption[Optional caption for list of figures]{\subref{fig:banditeval-senti} Evaluating the Bandit Framework on Subjectivity Dataset, \subref{fig:completeeval-senti} Evaluating the Complete Pipeline on Subjectivity Dataset [Decision trees as Predictive Model]}
\label{fig:subfigureExample}
\end{figure}
% bandit - Amazon reviews
% complete framework
\begin{figure}[ht]
\centering
\subfigure[]{
    \includegraphics[scale=0.10]{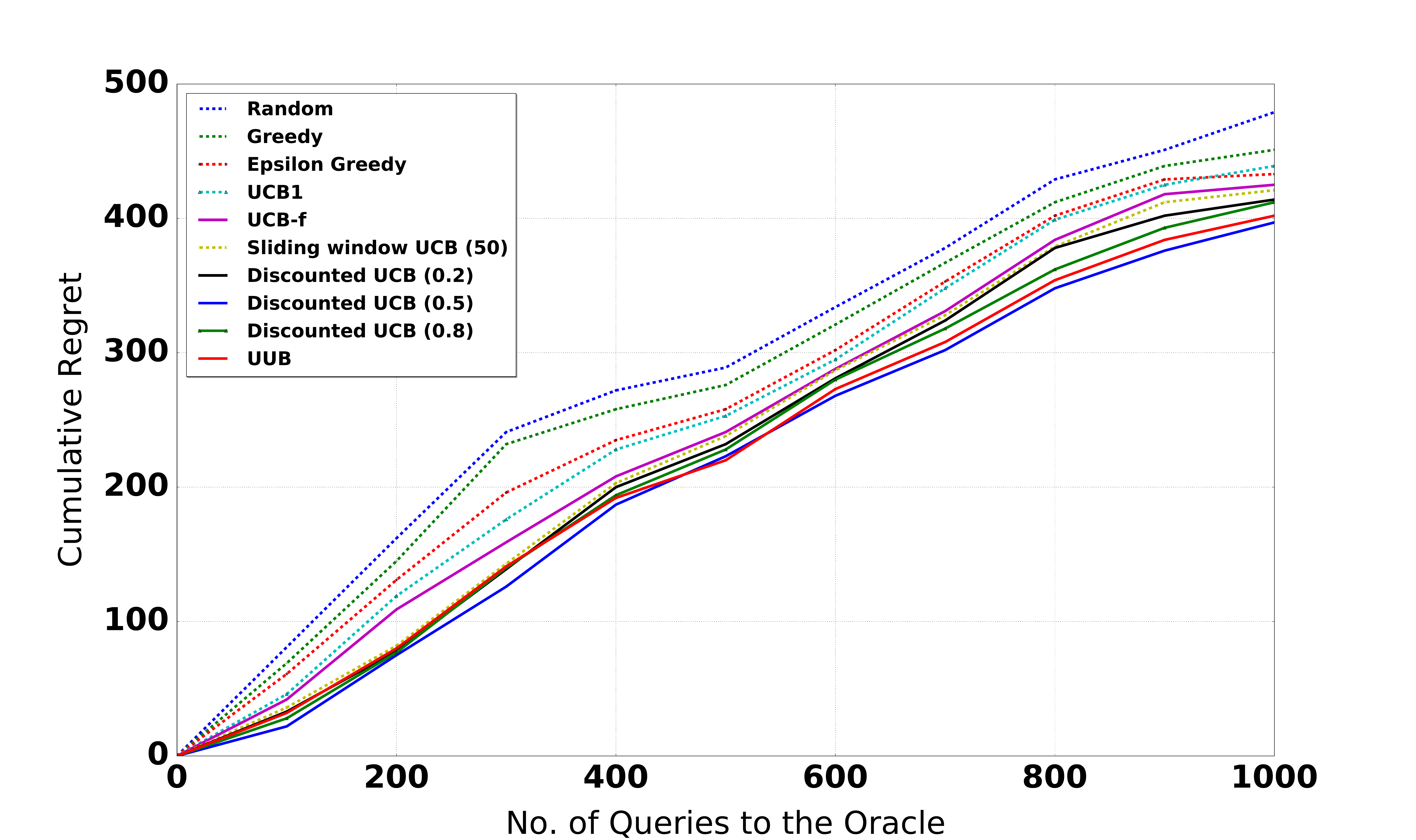}
    \label{fig:banditeval-subj}
}
\subfigure[]{
    \includegraphics[scale=0.10]{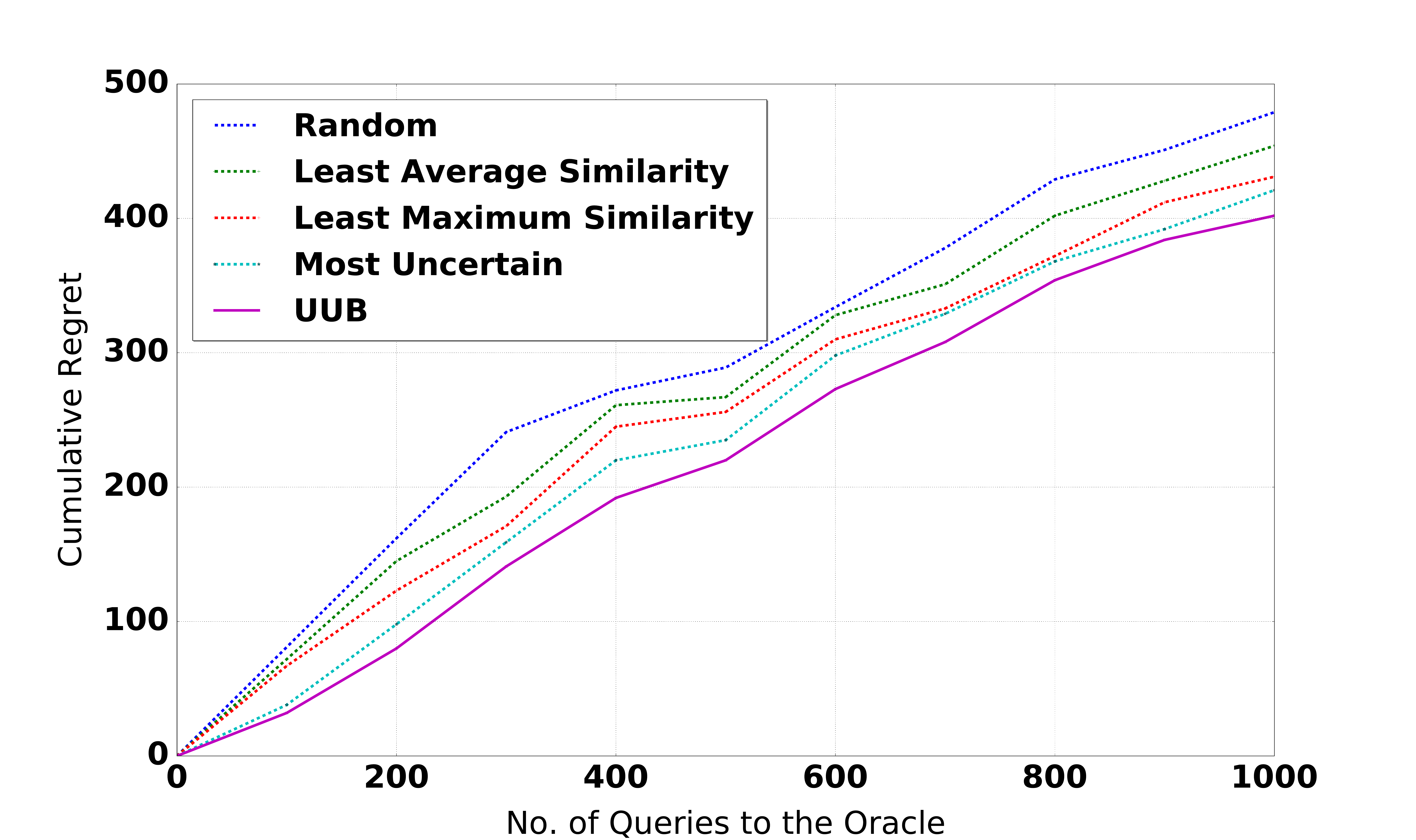}
    \label{fig:completeeval-subj}
}
\caption[Optional caption for list of figures]{\subref{fig:banditeval-senti} Evaluating the Bandit Framework on Amazon Reviews Dataset, \subref{fig:completeeval-senti} Evaluating the Complete Pipeline on Amazon Reviews Dataset [Decision trees as Predictive Model]}
\label{fig:subfigureExample}
\end{figure}

%\cleardoublepage
%\subsection{Results with Logistic Regression, SVMs, Random Forest, Neural Network}

\begin{figure}
\centering
\includegraphics[scale=0.10]{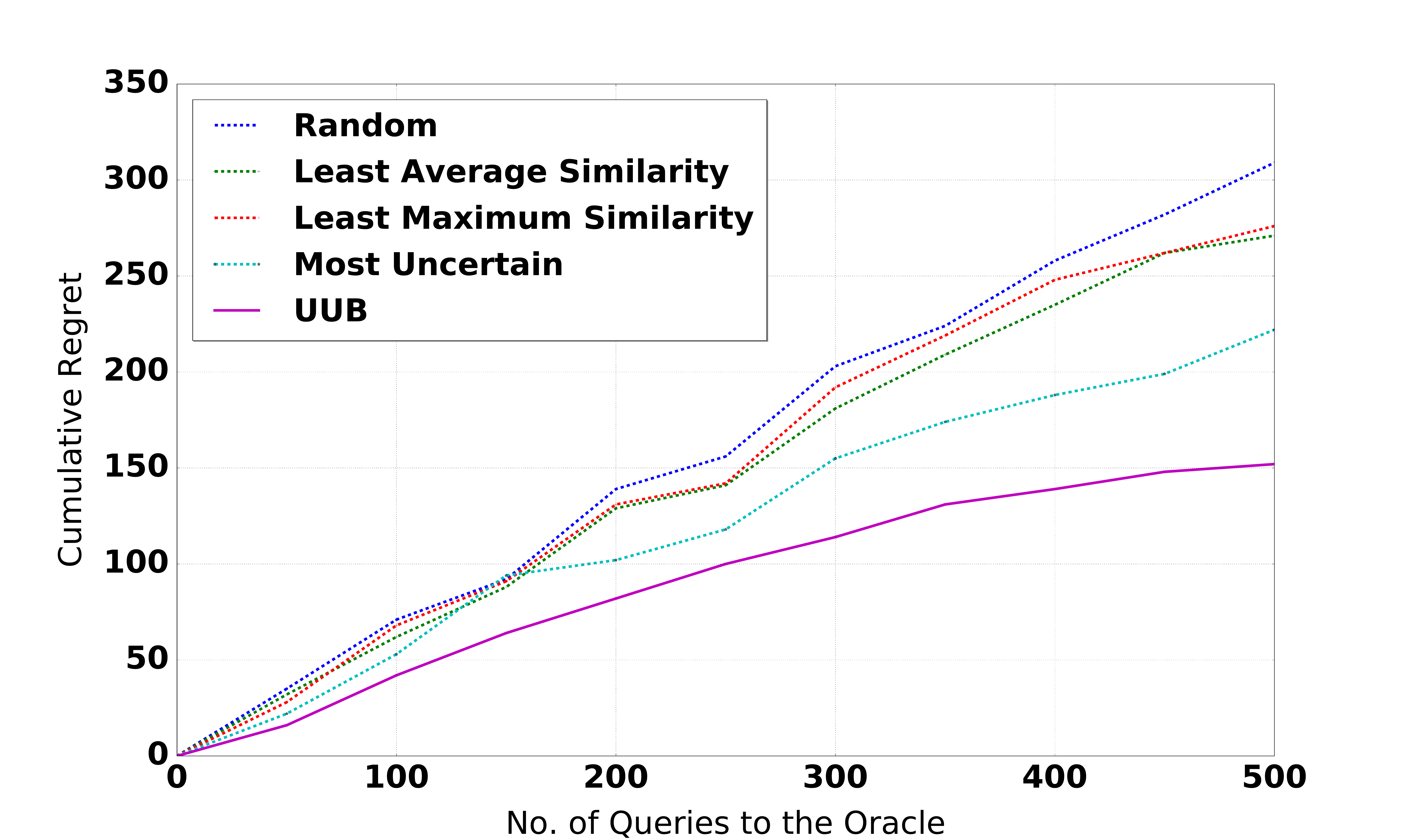}
\caption{Evaluating the Complete Pipeline on Image Dataset [Logistic Regression as Predictive Model]}
\label{fig:image-logit}
\end{figure}

\begin{figure}
\centering
\includegraphics[scale=0.10]{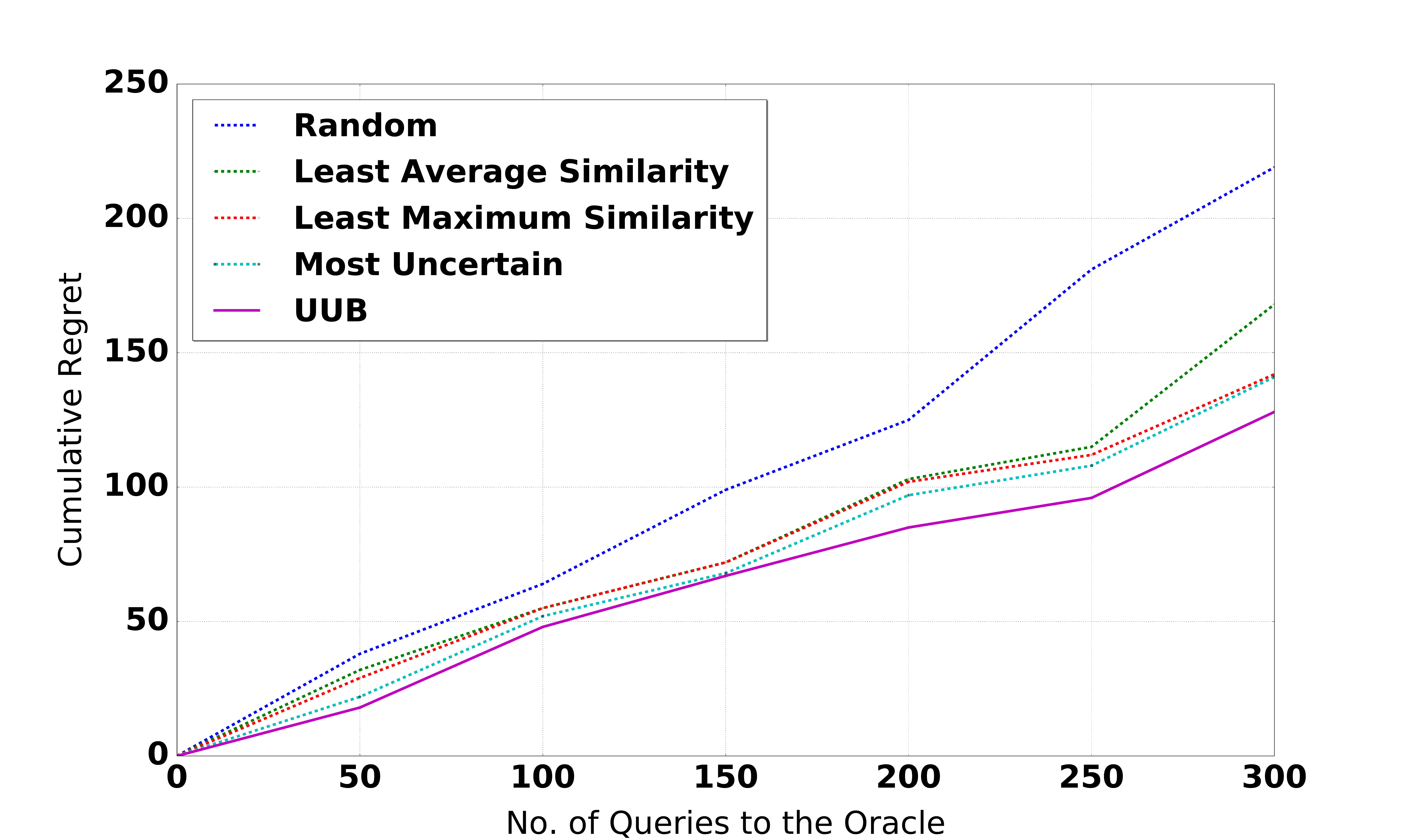}
\caption{Evaluating the Complete Pipeline on Subjectivity Dataset [Random Forests as Predictive Model]}
\label{fig:subj-RF}
\end{figure}

\begin{figure}
\centering
\includegraphics[scale=0.10]{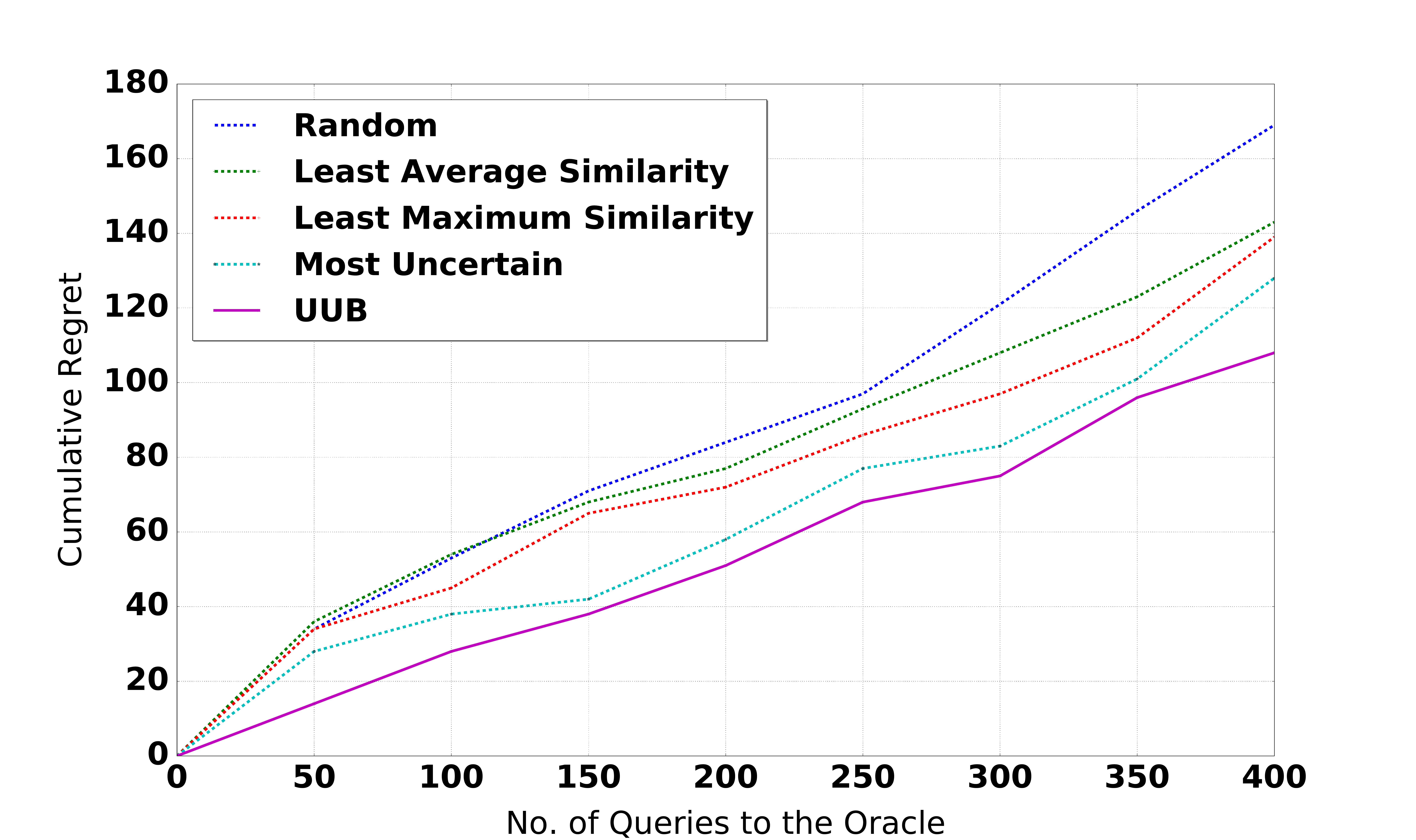}
\caption{Evaluating the Complete Pipeline on Sentiment Snippets [SVM as Predictive Model]}
\label{fig:senti-svm}
\end{figure}

\begin{figure}
\centering
\includegraphics[scale=0.10]{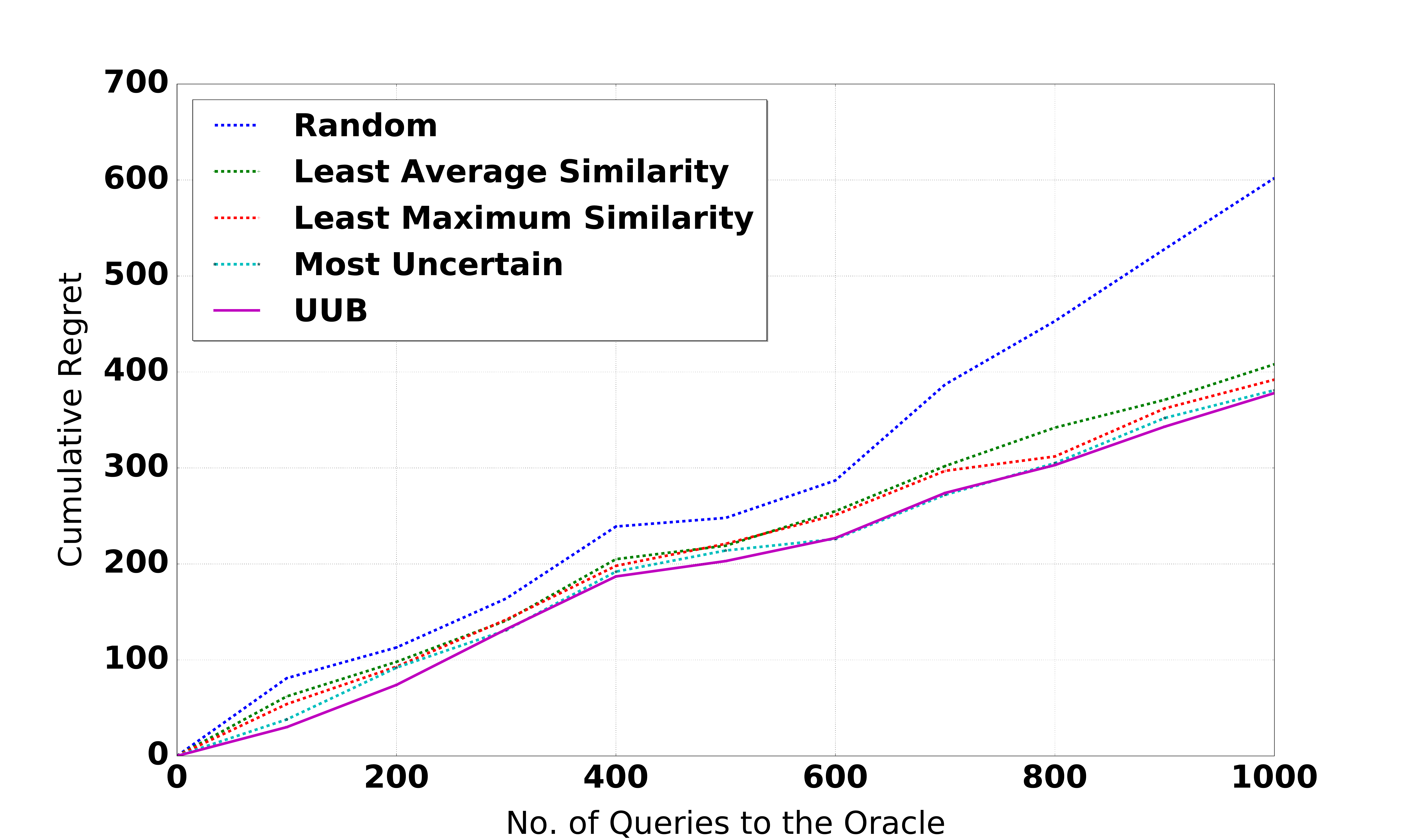}
\caption{Evaluating the Complete Pipeline on Amazon reviews [Neural Network (Multi layer perceptron with 5 hidden layers) as Predictive Model]}
\label{fig:senti-svm}
\end{figure}

%\clearpage
%\subsection{Retraining Models with Discovered Unknown Unknowns}
\begin{figure}
\centering
\includegraphics[scale=0.10]{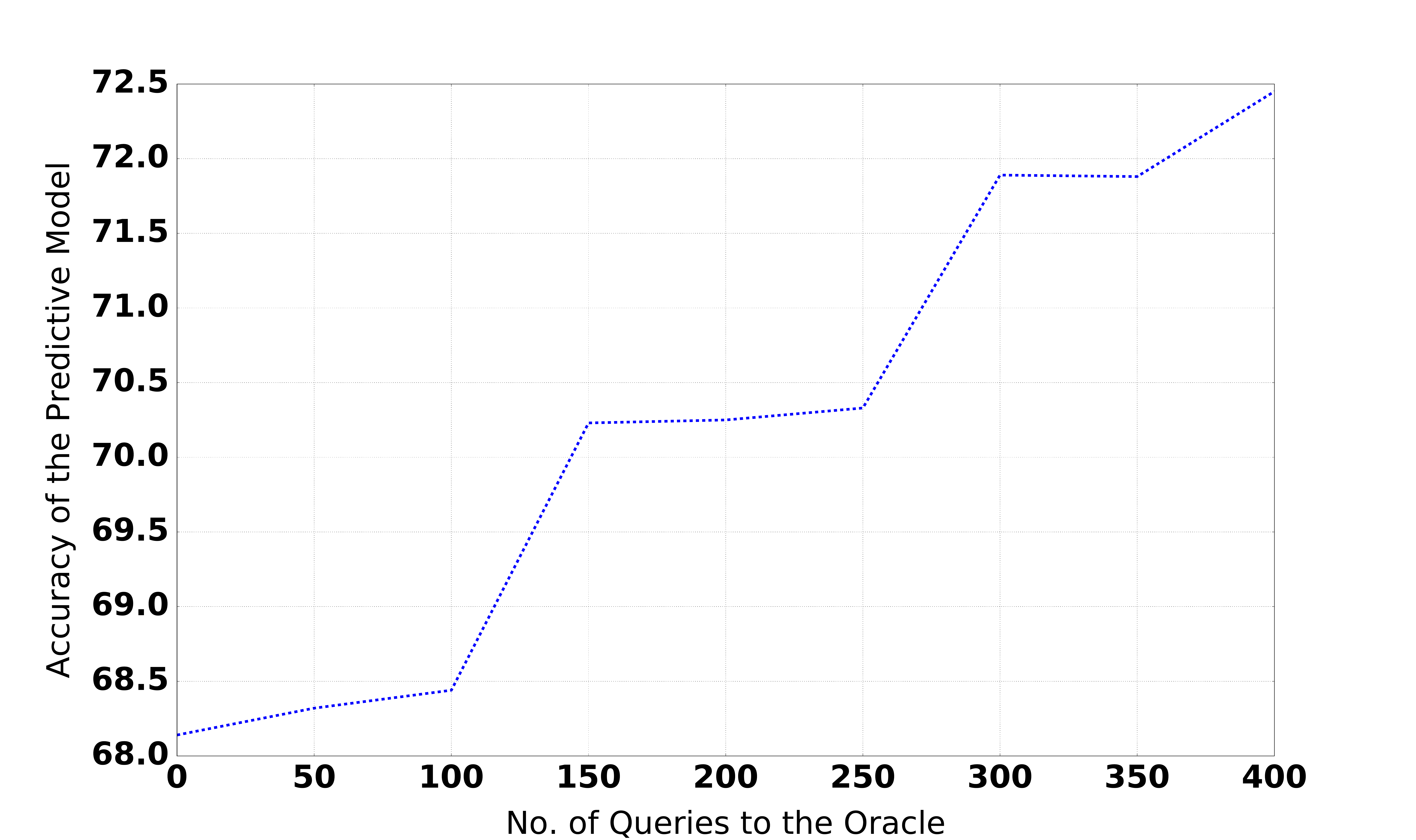}
\caption{Retraining the Predictive Model with Discovered Unknown Unknowns on Amazon reviews [Decision tree is the Predictive Model]}
\label{fig:appendix-learning}
\end{figure}
\end{document}